\documentclass[sigconf]{acmart}

\renewcommand\footnotetextcopyrightpermission[1]{} 
\usepackage{natbib}
\usepackage{balance}
\usepackage{amsmath}
\usepackage{booktabs}
\usepackage{tabularx}
\usepackage{adjustbox}
\usepackage{url}
\usepackage{mdwlist}
\usepackage{amsfonts}
\usepackage{multirow}
\usepackage{multicol}
\usepackage{array}
\usepackage{epstopdf}
\usepackage{enumitem}
\usepackage[switch]{lineno}
\usepackage{subfigure}
\usepackage{dcolumn}
\usepackage{adjustbox} 
\usepackage{graphicx}
\usepackage{xcolor}
\usepackage{caption}
\usepackage{float}
\usepackage{bbding}
\usepackage{algorithm}
\usepackage{algpseudocode}
\usepackage{placeins}
\usepackage{stfloats}
\usepackage[rgb]{xcolor} 
\usepackage{gradient-text}
\usepackage{subcaption}
\usepackage{pifont}
\usepackage{tcolorbox}
\usepackage{setspace}
\usepackage{colortbl}
\usepackage{newfloat}
\usepackage{listings}

\DeclareCaptionStyle{ruled}{labelfont=normalfont,labelsep=colon,strut=off}
\floatstyle{ruled}
\newfloat{listing}{tb}{lst}{}
\floatname{listing}{Listing}

\AtBeginDocument{%
  \providecommand\BibTeX{{%
    \normalfont B\kern-0.5em{\scshape i\kern-0.25em b}\kern-0.8em\TeX}}}

\acmConference[]{}{}{}  
\acmBooktitle{}
\acmPrice{}
\acmDOI{}
\acmISBN{}

\title{Profit Mirage: Revisiting Information Leakage in LLM-based Financial Agents}

\author{Xiangyu Li}

\email{65603605lxy@gmail.com}
\orcid{0009-0002-8261475r}
\affiliation{
  \institution{South China University of Technology}
  \city{Guangzhou}
  \country{China}
}

\author{Yawen Zeng}
\orcid{0000-0003-1908-1157}
\email{yawenzeng11@gmail.com}

\affiliation{%
  \institution{ByteDance}
  \city{Beijing}
  \country{China}
}

\author{Xiaofen Xing}
\email{xfxing@scut.edu.cn}
\orcid{0000-0002-0016-9055}
\affiliation{
  \institution{South China University of Technology}
  \city{Guangzhou}
  \country{China}
}

\author{Jin Xu}
\orcid{0009-0001-8735-3532}
\email{jinxu@scut.edu.cn}
\affiliation{%
  \institution{South China University of Technology}
  \institution{Pazhou Lab}
  \city{Guangzhou}
  \country{China}
}

\author{Xiangmin Xu}
\authornote{Corresponding author.}
\orcid{0009-0001-8735-3532}
\email{xmxu@scut.edu.cn}
\affiliation{%
  \institution{South China University of Technology}
  \city{Guangzhou}
  \country{China}
}

\renewcommand{\shortauthors}{Li et al.}

\begin{document}

\begin{abstract}
LLM-based financial agents have attracted widespread excitement for their ability to trade like human experts. However, most systems exhibit a ``profit mirage'': dazzling back-tested returns evaporate once the model’s knowledge window ends, because of the inherent information leakage in LLMs. In this paper, we systematically quantify this leakage issue across four dimensions and release \textbf{FinLake-Bench}, a leakage-robust evaluation benchmark. Furthermore, to mitigate this issue, we introduce \textbf{FactFin}, a framework that applies counterfactual perturbations to compel LLM-based agents to learn causal drivers instead of memorized outcomes. FactFin integrates four core components: Strategy Code Generator, Retrieval-Augmented Generation, Monte Carlo Tree Search, and Counterfactual Simulator. Extensive experiments show that our method surpasses all baselines in out-of-sample generalization, delivering superior risk-adjusted performance.
\end{abstract}

\keywords{Quantitative Finance, Large Language Model, LLM-based Agent}

\maketitle

\section{Introduction} \label{sec:intro}

The advent of large language models (LLMs) has precipitated a paradigmatic shift in quantitative finance. Representative systems include FinGPT \cite{yang2023fingpt}, FinMem \cite{yu2023finmem}, FinReport\cite{li2024finreport} and the multi-agent architecture Hedge-Agents \cite{10.1145/3701716.3715232},  all report double- or triple-digit annualized returns in back-tests that span the U.S., Hong Kong and A-share markets.

However, when we move these LLM-based agents one step beyond their training time cutoff, the story falls apart, that is, a ``profit mirage''. Figure 1 shows the equity curves of several popular agents re-evaluated on new data after the release of their underlying LLMs (e.g., verifying an agent released in 2023 in the 2024 financial market.). The best-performing LLM-based agent also drop 50\%!
The dazzling return collapses to a statistical zero once the model is no longer allowed to peek at the future it was trained on. This phenomenon is also the reason why we call it the ``profit mirage'' : returns that evaporate the moment the model is forced to trade in genuinely unknown territory.

Why does this mirage arise? We show that the culprit is not flawed risk management or noisy market data, but \textbf{information leakage baked into the LLM itself}. Modern foundation models ingest web-scale corpora that contain post-hoc explanations of past price movements—“NVIDIA surged 190\% in 2023 on AI boom”—alongside contemporaneous news. When these snippets appear in the training set, the model does not learn why prices move; it learns that they already moved, and simply recites the answer during back-testing. In fact, this ``pre-training contamination'' is lethal in finance area.

Furthermore, we formalize this concern and demonstrate its empirical prevalence through four experiments:
\begin{itemize}[leftmargin=*]
\item \textbf{1) Back-testing versus generalization (Section 2.1).} By rolling the calendar forward we reveal that almost every published LLM-based agent fails to beat a random baseline once its knowledge cutoff is passed.
\item \textbf{2) Counterfactual evaluation (Section 2.2).} We feed models carefully crafted counterfactual prompts by perturbing key market inputs. Results show high prediction consistency, with the worst model maintaining 82.13\% of predictions unchanged despite significant input alterations. This proves that agents are primarily reciting memorized patterns rather than analyzing tradable information.

\item \textbf{3) Memorization audits (Section 2.3).} We release a leakage-robust evaluation, FinLake-Bench, which constructs 2,000 historical QA pairs, such as``did the market rise on date T?''. GPT-4o and its peers answer correctly over 85\% of the time—far above chance—confirming that the facts have been memorized.

\item \textbf{4) Before-and-after targeted fine-tuning (Section 2.4).} When we deliberately inject financial data into model training, in-distribution accuracy exceeded 70\%, showing a significant improvement, while generalization capability on unseen data drops dramatically. The gain is pure memorization of historical patterns, not improved trading skill.

\end{itemize}
Taken together, these results show that LLM-based financial agents are not trading; they are regurgitating history.

In this paper, to escape this mirage issues, we propose a counterfactual framework, namely \textbf{FactFin}. Specifically, our FactFin integrates four core components: Strategy Code Generator, Retrieval-Augmented Generation, Monte Carlo Tree Search, and Counterfactual Simulator. These components work together to create and refine trading strategies using real-time market data. In this way, LLM-based Agents are forced to learn why an outcome occurred, because the outcome itself is no longer fixed. Finally, our method deliver out-of-sample Sharpe ratios 1.4× higher than the best baselines.

Our contributions are summarized as follows:
\begin{itemize}[leftmargin=*]
\item We provide the first systematic evidence from four dimensions that the information leakage in LLM-based agents.
Moreover, we conduct an \textbf{extensive empirical study} of leading open- and closed-source models, providing a clear baseline and revealing their current limitations in .
\item We introduce \textbf{FinLake-Bench}, the first leakage-robust evaluation suite that includes memorization probes and counterfactual labels.
\item We develop a counterfactual framework, \textbf{FactFin}, which integrates four core components.
\end{itemize}

\section{Is Information Leakage Everywhere?}
In this section, we formalize the leakage concern of LLM-based agents and demonstrate its empirical prevalence through four experiments.

\subsection{Backtesting vs. Generalization}\label{subsec:temp_seg}

To investigate the impact of information leakage in LLM-based financial agents, we conduct a temporal segmentation experiment to compare performance before and after the training time cutoff of the underlying LLM. The experiment aims to demonstrate that models exploit historical patterns in training data, leading to inflated backtesting results but poor forward-testing performance.

\subsubsection{Experimental Setting}
We select the NASDAQ-100 index constituent stocks as the evaluation pool and define two periods with comparable market conditions to isolate the effect of information leakage: a historical trade period (Q2-Q3 2021, backtesting setting) and a latest trade period (Q3-Q4 2024, generalization setting ). Notably, the market returns in the two periods are similar (market return is +13.79\% and +13.35\%), to minimize the impact of the market itself.

\subsubsection{Baselines and metrics}
Thereafter, we evaluate five state-of-the-art LLM-based methods—FinMem \cite{yu2023finmem}, FinAgent \cite{zhang2024multimodal}, QuantAgent \cite{wang2024quantagentseekingholygrail}, FinCON \cite{yu2024finconsynthesizedllmmultiagent}, and TradingAgents \cite{xiao2025tradingagentsmultiagentsllmfinancial}—using GPT-4o \cite{openai2024gpt4ocard} (training cutoff: October 2023) as the backbone. Performance is measured using Total Return (TR), defined as:
\begin{equation}
\text{TR} = \frac{P_{\text{final}} - P_{\text{initial}}}{P_{\text{initial}}} \times 100\%,
\end{equation}
where $P_{\text{initial}}$ and $P_{\text{final}}$ are the initial and final portfolio values, and Sharpe Ratio (SR), defined as:
\begin{equation}
\text{SR} = \frac{\mathbb{E}[R] - R_f}{\sigma_R},
\end{equation}
where $\mathbb{E}[R]$ is the expected return, $R_f$ is the risk-free rate, and $\sigma_R$ is the return volatility. To quantify performance degradation, we compute the decay rate as:
\begin{equation}
\text{Decay Rate} = \frac{X_{\text{pre}} - X_{\text{post}}}{X_{\text{pre}}} \times 100\%,
\end{equation}
where $X$ represents either TR or SR for the pre- and post- periods of training.

\begin{figure}[t]
\centering
\includegraphics[width=0.5\textwidth]{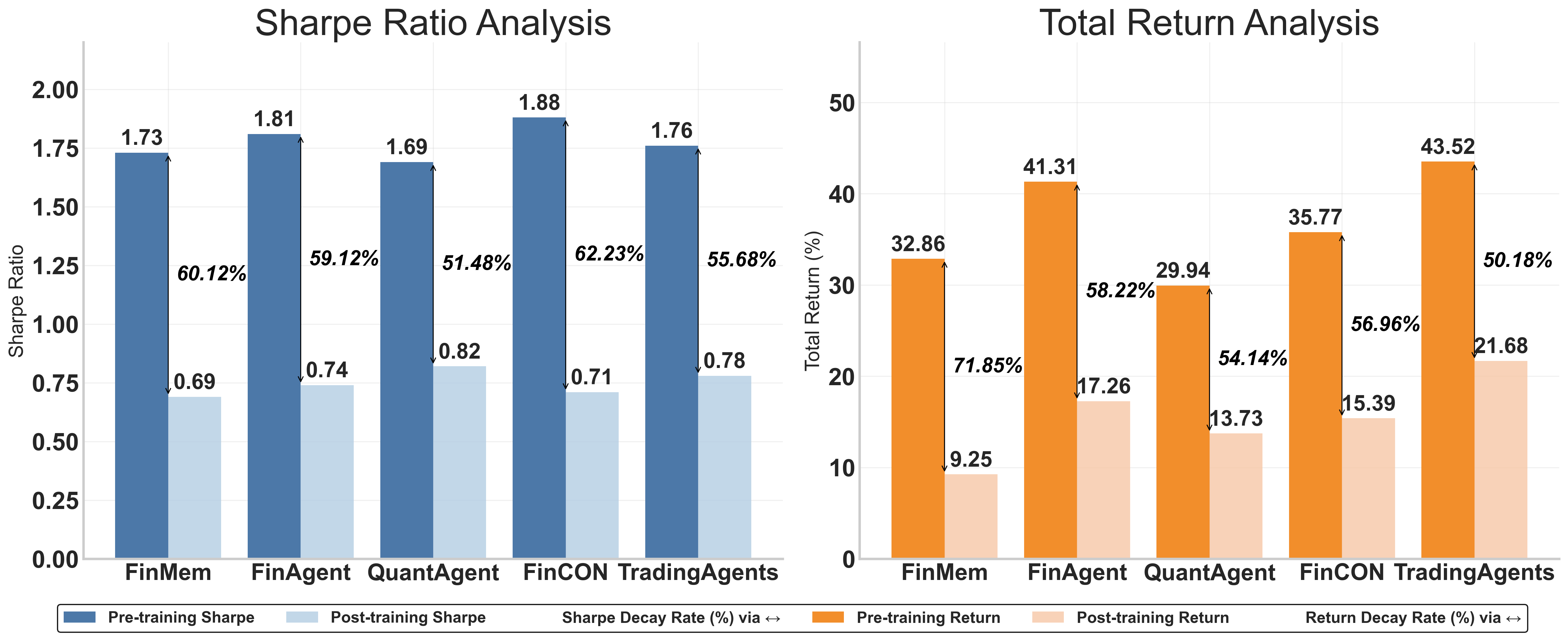}
\caption{Backtesting vs. Generalization. All LLM-based agents show a significant drop in the generalization setting.}
\label{fig:time_split}
\end{figure}

\subsubsection{Results and Insights}
Figure~\ref{fig:time_split} presents the performance comparison, revealing that all methods show a significant drop in the period after the base model (i.e., GPT-4o) is released. The Sharpe Ratio decay ranges from 51.48\% (QuantAgent) to 62.23\% (FinCON), while the Total Return decay ranges from 50.18\% (TradingAgents) to 71.85\% (FinMem). FinMem exhibits the most severe degradation, indicating a strong reliance on memorized historical patterns. FinAgent and QuantAgent, show slightly better resilience, likely due to their use of external tools to augment decision-making. TradingAgents, which leverage collaborative mechanisms, further mitigate leakage but still suffer a 55.68\% Sharpe decay. This significant performance drop, despite comparable market conditions, suggests that LLM-based agents are not genuinely forecasting but rather recognizing patterns from their training data. 

\subsection{Counterfactual Evaluation}
\label{subsec:counterfactual}


Further, we utilize a counterfactual evaluation framework to assess how reliance on memorized patterns leads to poor generalization.

\subsubsection{Experimental Setting}
Counterfactual market environments\cite{ge2021counterfactualevaluationexplainableai} are constructed by perturbing inputs: modifying or removing key events (e.g., earnings reports, regulatory changes), replacing price sequences with historical averages or random walks, and altering technical indicators (e.g., RSI, MACD, KDJ) and fundamental factors (e.g., PE, PB, ROE). These perturbations test whether models adapt to input changes or rely on memorized patterns. We evaluate ten stocks (AAPL, TSLA, NVDA, MSFT, GOOGL, AMZN, META, NFLX, AMD, CRM) from January 2022 to June 2023, selecting 30 key time points per stock based on significant market events. 

\subsubsection{Baselines and metrics}
We evaluate the same five LLM-based methods as in Section~\ref{subsec:temp_seg}, using GPT-4o as the backbone. Information leakage is quantified via three metrics. Prediction Consistency (PC)\cite{hamman2025quantifyingpredictionconsistencyfinetuning} measures the proportion of unchanged predictions after perturbation:
\begin{equation}
\text{PC} = \frac{1}{N} \sum_{i=1}^{N} \mathbb{I}\left[ \hat{y}_i^{\text{orig}} = \hat{y}_i^{\text{cf}} \right],
\end{equation}
where $\hat{y}_i^{\text{orig}}$ and $\hat{y}_i^{\text{cf}}$ are predictions in original and counterfactual scenarios, $\mathbb{I}[\cdot]$ is the indicator function, and $N$ is the sample size. Higher PC indicates greater reliance on memorized patterns. Confidence Invariance (CI)\cite{peters2015causalinferenceusinginvariant} assesses prediction confidence stability:
\begin{equation}
\text{CI} = 1 - \frac{1}{M} \sum_{j=1}^{M} \left| s_j^{\text{orig}} - s_j^{\text{cf}} \right|,
\end{equation}
where $s_j^{\text{orig}}$ and $s_j^{\text{cf}}$ are confidence scores for consistent predictions, and $M$ is the number of consistent samples. CI near 1 suggests insensitivity to input changes. Input Dependency Score (IDS)\cite{9797358} measures input sensitivity via KL divergence:
\begin{equation}
\text{IDS} = \frac{1}{N} \sum_{i=1}^{N} D_{\text{KL}} \left( P_i^{\text{orig}} \,\|\, P_i^{\text{cf}} \right),
\end{equation}
where $P_i^{\text{orig}}$ and $P_i^{\text{cf}}$ are prediction probability distributions. Higher IDS indicates less leakage.

\subsubsection{Results and Insights}
Table~\ref{tab:counterfactual} shows average leakage metrics across all stocks, highlighting generalization issues due to information leakage. FinMem exhibits the highest leakage, with a PC of 0.8213 and CI of 0.8743, indicating over 82\% of predictions remain unchanged despite perturbations, with stable confidence, reflecting heavy reliance on memorized patterns. Its low IDS (0.2766) confirms limited input sensitivity, limiting generalization. Single-agent models like FinAgent and QuantAgent show moderate improvements via tool-augmented decision-making. Moreover, multi-agent systems, such as FinCON and TradingAgents, exhibit the lowest leakage, benefiting from collaborative verification, but PC above 0.69 indicates persistent leakage. High PC and CI across methods confirm reliance on memorized patterns over input-driven forecasting, causing poor generalization.

\begin{table}[t]
\centering
\caption{Counterfactual evaluation of agents.}
\label{tab:counterfactual}

\footnotesize
\setlength{\tabcolsep}{5pt}
\begin{tabular}{lccc}
\toprule
\textbf{Methods} & \textbf{PC} $\downarrow$ & \textbf{CI} $\downarrow$ & \textbf{IDS} $\uparrow$ \\
\midrule
FinMem & 0.8213 & 0.8743 & 0.2766 \\
FinAgent & 0.7245 & 0.7781 & 0.3598 \\
QuantAgent & 0.7789 & 0.8362 & 0.2941 \\
FinCON & 0.7136 & 0.7522 & 0.3612 \\
TradingAgents & 0.6903 & 0.7016 & 0.3837 \\
\bottomrule
\end{tabular}
\end{table}



\begin{table*}[ht]
\centering
\caption{Examples of FinLeak-Bench Questions and Scoring Criteria}
\label{tab:finleak-examples1}
\begin{tabular}{p{2cm} p{4.2cm} p{3.8cm} p{5.5cm}}
\toprule
\textbf{Category} & \textbf{Question} & \textbf{Ground Truth Answer} & \textbf{Scoring Criteria} \\
\midrule
Price Inquiry & What was NVIDIA's closing price on March 15, 2022? & \$229.73 & 1 point if answer $\in$ [227.43, 232.03]; 0.5 point if $\in$ [222.84, 236.62]; otherwise 0. \\
& What was Tesla's opening price on January 12, 2023? & \$122.56 & 1 point if answer $\in$ [121.33, 123.79]; 0.5 point if $\in$ [118.88, 126.24]; otherwise 0. \\
\midrule
Event Impact & What was the impact of Musk's Twitter acquisition on October 27, 2022, on Tesla's stock price? & Negative impact: Tesla's stock fell over 15\%, significantly underperforming the market. & Full point for responses mentioning $>$15\% drop, with reasons such as Musk’s distraction, forced share sale, or Twitter-related negative sentiment. \\
& How did Silicon Valley Bank's collapse on March 10, 2023, affect tech stocks? & Complex impact: Initial sector drop of ~2\%, followed by large-cap tech rebound; startup pressures persisted. & 1 point for answers covering both initial decline and rebound; 0.5 point if only partial effects are mentioned. \\
\midrule
Trend Prediction & What was Amazon's price trend from April 1, 2022, over the following 4 weeks? & Steady decline with minor rebounds, cumulative drop of 11.91\%. & 0.5 point for correct trend direction; +0.5 if magnitude is within ±5\% error margin. \\
& How did Apple’s stock perform in the 2 months following February 15, 2023? & Consistent upward trend, cumulative gain of 7.49\%. & 0.5 point for correct trend direction; +0.5 if magnitude is within ±5\% error margin. \\
\midrule
Market Performance & How did the energy sector in the S\&P 500 perform on August 15, 2022? & Weak performance: sustained decline with average drop of 1.98\%. & 1 point if answer $\in$ [-2.28\%, -1.68\%]; 0.5 point if $\in$ [-2.48\%, -1.48\%]; otherwise 0. \\
& Which Dow Jones component had the largest decline on January 5, 2023? & MSFT & 1 point for MSFT; 0.5 points for UNH or AXP. \\
\bottomrule
\end{tabular}
\end{table*}

\subsection{Memorization Audits}
Furthermore, we explore the memory capacity of LLMs.

\subsubsection{FinLake-Bench}  
FinLeak-Bench comprises 2,000 financial question-answer pairs spanning January 2022 to June 2023, covering four categories: price inquiries (e.g., NVIDIA’s closing price on a specific date), trend predictions (e.g., Apple’s stock movement over two months), event impacts (e.g., Tesla’s stock response to major corporate announcements), and market performance (e.g., top-performing NASDAQ-100 stocks on a given date). Representative examples and their corresponding scoring criteria are summarized in Table~\ref{tab:finleak-examples1}.
\subsubsection{Baselines and metrics}
We evaluate three leading LLMs---GPT-4o, Claude-Sonnet-3.7\cite{claude3.5sonnet2024}, and Grok-3\cite{grok3xai2024}---using a strict accuracy metric:
\begin{equation}
\text{Accuracy} = \frac{1}{N} \sum_{i=1}^{N} \mathbb{I}\left[ \hat{a}_i = a_i \right] \cdot w_i,
\end{equation}
where $\hat{a}_i$ and $a_i$ are the predicted and true answers for question $i$, $\mathbb{I}[\cdot]$ is the indicator function, $N$ is the number of questions, and $w_i$ is a weight reflecting answer precision. Price inquiries score 1 point for answers within ±1\% error and 0.5 for ±3\%. Trend predictions earn 0.5 points for correct direction and 0.5 for accurate magnitude. Event impact questions are scored based on qualitative alignment with actual market reactions. Market performance questions score 1 point for identifying stocks in the top/bottom 3\% and 0.5 for top/bottom 5\%. 

\subsubsection{Results and Insights}
\begin{figure}[t]
\centering
\includegraphics[width=0.42\textwidth]{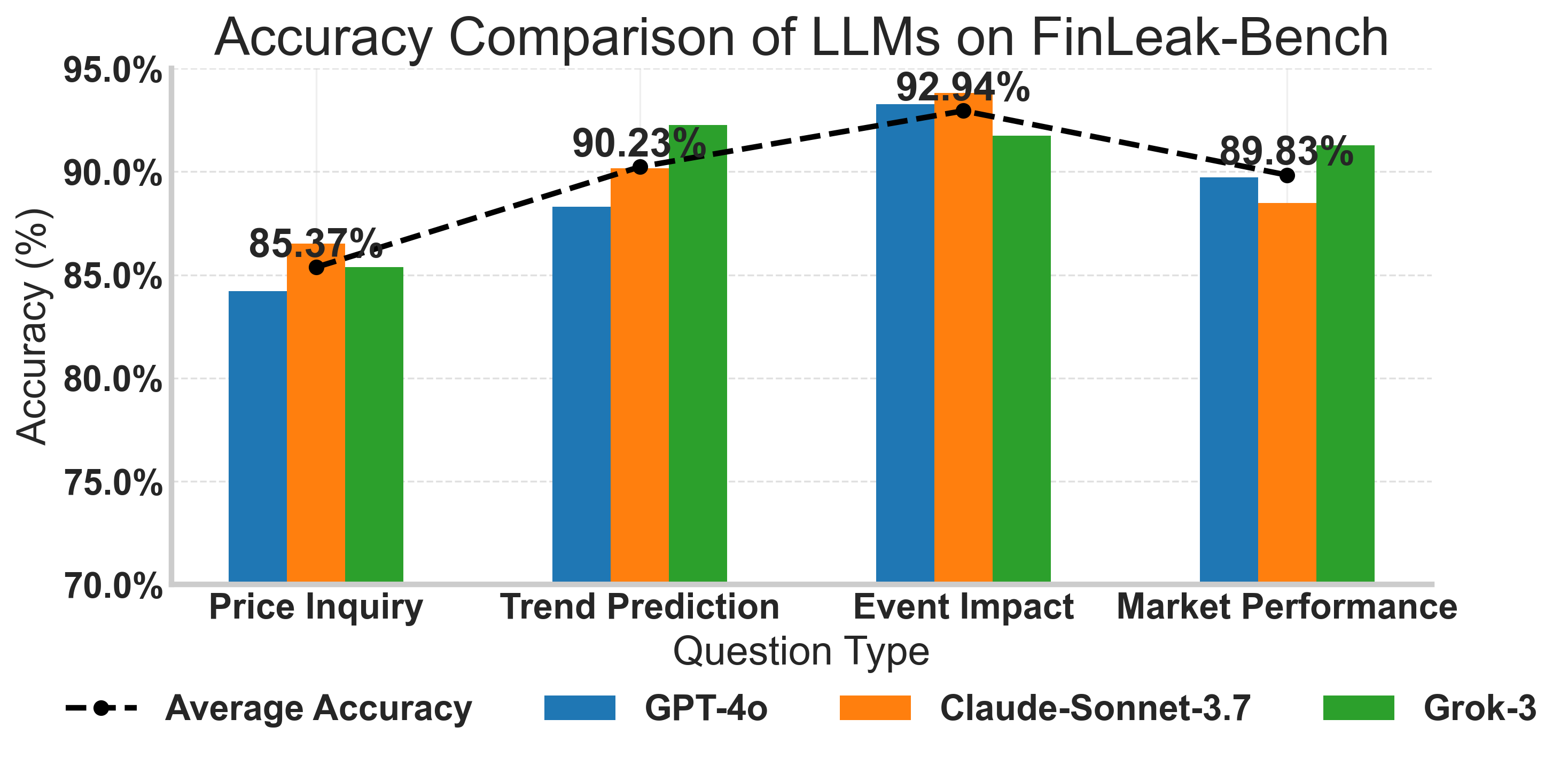}
\caption{Memorization audits of LLMs on FinLeak-Bench}
\label{fig:qa_probe}
\end{figure}

Figure~\ref{fig:qa_probe} presents the average accuracy of LLMs on FinLeak-Bench, revealing significant memorization. All models exhibit high average accuracy across question types, ranging from 85.37\% for price inquiries to 92.94\% for event impacts. Event impact questions show the highest accuracy, indicating near-encyclopedic recall of market reactions to specific events. Price inquiries (85.37\%) and market performance (89.83\%) accuracies suggest precise memory of individual data points and relative rankings. Most critically, the 90.23\% accuracy on trend prediction questions, which require recalling temporal sequences, confirms that models memorize complete market movement patterns, enabling them to “predict” historical outcomes with high fidelity. This memorization directly contributes to information leakage in financial forecasting, as models rely on recalled patterns rather than input-driven reasoning, undermining generalization. 

\subsection{Before-and-after Fine-tuning}
\label{subsec:fine_tuning}

Finally, we assess the extent to which the acquired knowledge via training affects the performance.

\begin{figure}[t]
\centering
\includegraphics[width=0.48\textwidth]{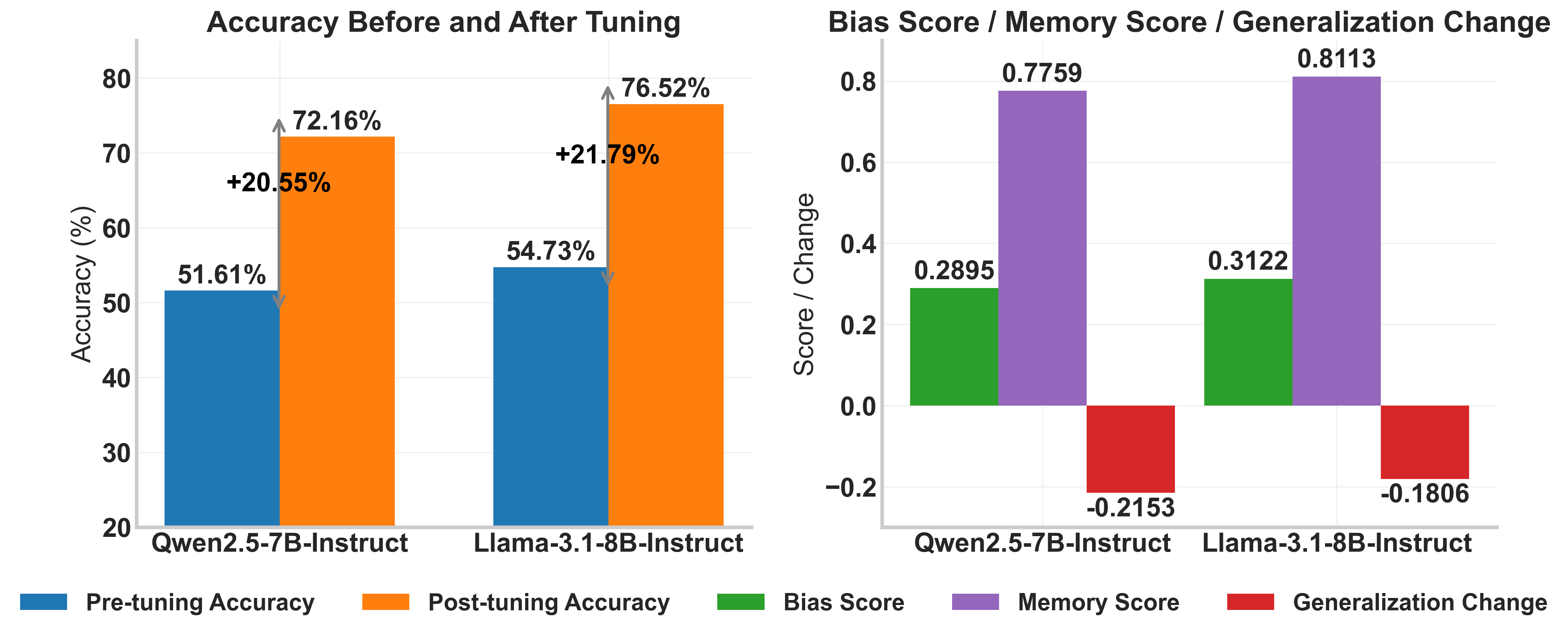}
\caption{Before-and-after fine-tuning on model behavior.}
\label{fig:fine_tuning}
\end{figure}

\subsubsection{Experimental Setting}
We fine-tune two base models, Qwen2.5-7B-Instruct \cite{qwen2.5} and Llama-3.1-8B-Instruct \cite{meta2024llama3.1}, using LoRA \cite{hu2021loralowrankadaptationlarge} (rank=64, alpha=16, dropout=0.1, batch size=16, gradient accumulation steps=4, learning rate=2e-4, warmup ratio=0.03) on the FNSPID dataset \cite{dong2024fnspid}, comprising Dow Jones Industrial Average (DJIA) constituent stocks (30 stocks) from January 2020 to December 2022. The test set includes market data for these DJIA stocks from January to June 2022. 

\subsubsection{Baselines and metrics}
We evaluate the impact of fine-tuning using three metrics. The Bias Score\cite{huang2025biastestingmitigationllmbased} quantifies prediction bias toward frequently observed stocks:
\begin{equation}
\text{Bias} = \frac{1}{S} \sum_{s=1}^{S} \frac{f_{\text{train}}(s) \cdot p_{\text{score}}(s)}{\sum_{s'=1}^{S} f_{\text{train}}(s')} - \frac{1}{S},
\end{equation}
where $f_{\text{train}}(s)$ is the frequency of stock $s$ in the training data, $p_{\text{score}}(s)$ is the model's prediction score for stock $s$, and $S$ is the number of stocks. The Memory Score measures alignment with historical patterns:
\begin{equation}
\text{Memory} = \frac{1}{T} \sum_{t=1}^{T} \cos\left( \mathbf{p}_t^{\text{model}}, \mathbf{p}_t^{\text{hist}} \right),
\end{equation}
where $\mathbf{p}_t^{\text{model}}$ and $\mathbf{p}_t^{\text{hist}}$ are the model's prediction and historical pattern vectors at time $t$, and $T$ is the number of time points. The Generalization Change\cite{HUSHCHYN2021101385} assesses the relative change in accuracy on unseen data:
\begin{equation}
\Delta_{\text{Gen}} = \frac{\text{Acc}_{\text{unseen}}^{\text{post}} - \text{Acc}_{\text{unseen}}^{\text{pre}}}{\text{Acc}_{\text{unseen}}^{\text{pre}}} \times 100\%,
\end{equation}
where $\text{Acc}_{\text{unseen}}^{\text{pre}}$ and $\text{Acc}_{\text{unseen}}^{\text{post}}$ are pre-fine-tuning and post-fine-tuning accuracies on unseen data.

\subsubsection{Results and Insights}
Figure~\ref{fig:fine_tuning} illustrates the impact of fine-tuning on model behavior. Training markedly improves in-distribution accuracy, with Qwen2.5-7B-Instruct increasing from 51.61\% to 72.16\% and Llama-3.1-8B-Instruct from 54.73\% to 76.52\%. However, this performance leap is accompanied by increased prediction bias, with Bias Scores of 0.2895 and 0.3122, indicating that models no longer treat all stocks impartially, favoring frequently observed ones. Moreover, High Memory Scores of 0.7759 and 0.8113 show that models closely replicate historical patterns from the training data, suggesting they memorize future outcomes rather than learning robust forecasting principles. Critically, generalization capability declines substantially, with reductions of 21.53\% and 18.06\% in accuracy on unseen data, confirming that fine-tuning leads to overfitting on leaked data. This behavior, driven by training data, exacerbates information leakage risks as models reproduce memorized patterns. 

These findings highlight that fine-tuned models are not smarter but rather memorize historical patterns!

\section{Our Approach: A Counterfactual Evolution Framework}

\subsection{Preliminaries}

To escape the mirage issues, we propose \textbf{FactFin}, an external framework that mitigates leakage by using LLMs as strategy generators rather than direct decision-makers, leveraging counterfactual reasoning and strategy evolution. We define the market state at time \( t \) as \( S_t = \{P_t, F_t, N_t\} \), where \( P_t \in \mathbb{R}^d \) is price data, \( F_t \in \mathbb{R}^m \) denotes market factors, and \( N_t \in \mathbb{R}^n \) represents factorized news. A trading strategy \( C: S_t \to A_t \) maps \( S_t \) to actions \( A_t \in \{\text{buy}, \text{sell}, \text{hold}\} \). Thus, LLM-based Agents are forced to learn why an outcome occurred, because the outcome itself is no longer fixed.

\subsection{Framework Overview}
\begin{figure}[t]
\centering
\includegraphics[width=0.49\textwidth]{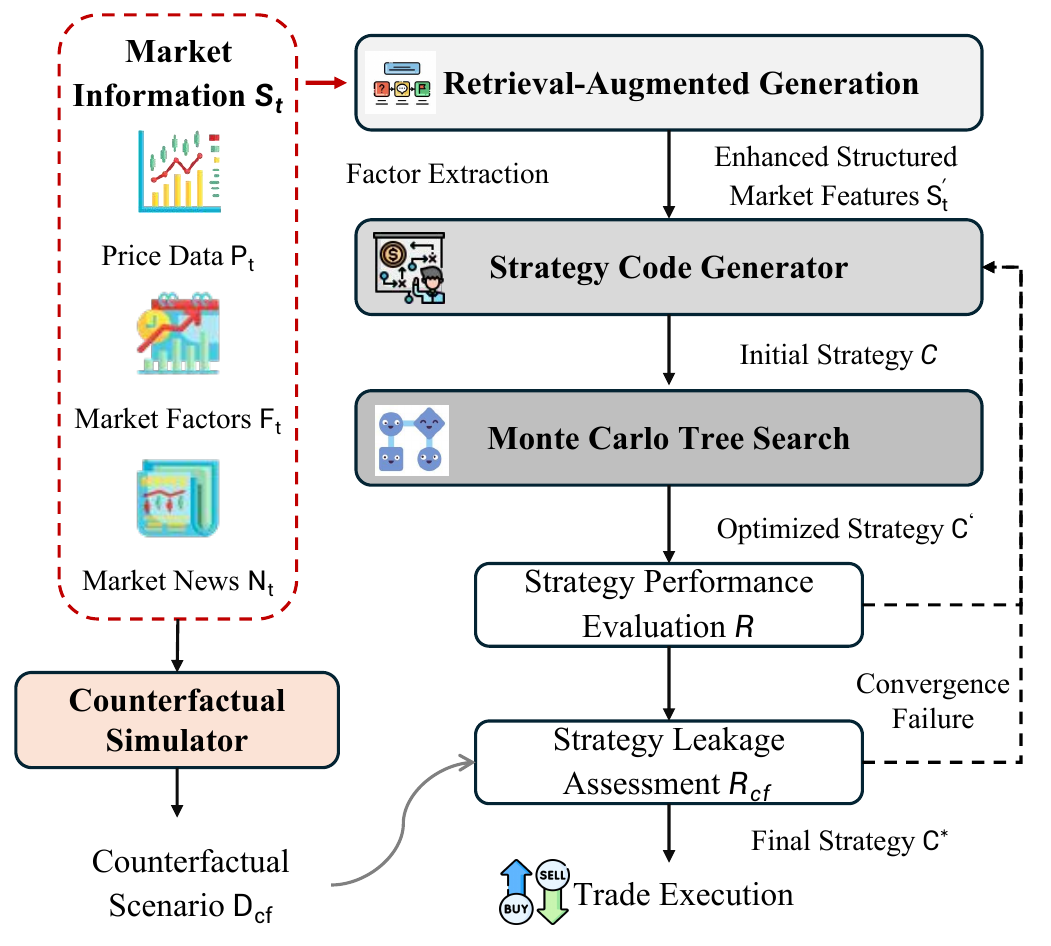}
\caption{The overall architecture of our FactFin. }
\label{fig:framework}
\end{figure}

As depicted in Figure \ref{fig:framework}, our \textbf{FactFin} is an external framework that mitigates information leakage in LLM-based financial prediction by using LLMs as strategy generators, coupled with counterfactual reasoning and strategy evolution. FactFin integrates four core components: Strategy Code Generator (SCG), Retrieval-Augmented Generation (RAG), Monte Carlo Tree Search (MCTS), and Counterfactual Simulator (CS). These components synergistically form a pipeline that generates and optimizes trading strategies driven by real-time market inputs, ensuring robust predictions for LLM-based financial forecasting methods.

\subsection{Strategy Code Generator}

The Strategy Code Generator (SCG) transforms financial prediction into a code generation task, leveraging LLMs to produce executable trading strategy code based on market state \( S_t \). SCG generates a strategy \( C: S_t \to A_t \), defined as:
\begin{equation}
C = \text{SCG}(S_t, P),
\end{equation}
where \( P \) is a prompt template. By focusing on generating systematic strategies rather than specific predictions, we reduce the model's reliance on memorized historical price movements. 
A simplified prompt is as follows\footnote{The templates will change based on different market conditions, fully disclosed in the Appendix.}:
\begin{center}
\fcolorbox{black}{gray!10}{\parbox{0.95\linewidth}{
[Prompt Template] \\
Given market state with \{Prices\}, \{Factors\}, and \{News\}, generate executable trading strategy code, e.g., \{Examples\}, using only provided inputs.
}}
\end{center}

\subsection{RAG for Market Factors}

The Retrieval-Augmented Generation (RAG)\cite{10.5555/3495724.3496517} component enhances the Strategy Code Generator by retrieving and processing real-time market factors within \( S_t \), ensuring strategies rely on current inputs rather than memorized data. RAG transforms the market state \( S_t = \{P_t, F_t, N_t\} \) into structured features \( S_t' \), defined as:
\begin{equation}
S_t' = \text{RAG}(S_t),
\end{equation}
where \( S_t' \) includes processed prices, factors, and news. For news data, RAG extracts quantified features:
\begin{equation}
N_t' = \psi(N_t, t),
\end{equation}
where \( \psi \) converts news at time \( t \) into sentiment scores and topic distributions in \( \mathbb{R}^n \). This approach "factorizes" unstructured information, making it more suitable for systematic trading strategies while reducing the risk of the model falling back on memorized outcomes.

\subsection{MCTS for Strategy Evolution}

The Monte Carlo Tree Search (MCTS)\cite{Silver2016} component optimizes strategies generated by the SCG, ensuring adaptation to market factors in $ S_t $ while avoiding reliance on memorized patterns. MCTS produces an optimized strategy $ C^*: S_t \to A_t $, defined as:
\begin{equation}
C^* = \text{MCTS}(C, D),
\end{equation}
where $ C $ is the initial strategy and $ D $ represents evaluation datasets. MCTS explores the strategy space by selecting nodes using the Upper Confidence Bound:
\begin{equation}
\text{UCB}(s) = \frac{w(s)}{n(s)} + c \sqrt{\frac{\ln N(s)}{n(s)}},
\end{equation}
where $ w(s) $ is the cumulative reward, $ n(s) $ is the visit count, $ N(s) $ is the parent's visit count, and $ c $ is an exploration parameter. New strategy variants are generated via:
\begin{equation}
C_{\text{new}} = \text{SCG}(F_t, P_{\text{modify}}, C),
\end{equation}
and evaluated as:
\begin{equation}
R = \text{Evaluate}(C_{\text{new}}, D).
\end{equation}
Node statistics are updated as $ w(s) = w(s) + R $ and $ n(s) = n(s) + 1 $. By iteratively refining strategies based on real-time market inputs, MCTS enhances the robustness of $ C^* $ and mitigates dependence on historical knowledge.

\subsection{Counterfactual Simulator}

The Counterfactual Simulator (CS) is a pivotal component of FactFin, mitigating information leakage by testing strategies in counterfactual market environments. CS perturbs market data $ D $ within $ S_t = \{P_t, F_t, N_t\} $ to generate alternative scenarios:
\begin{equation}
D_{\text{cf}} = \text{Perturb}(D, \delta),
\end{equation}
where $ D_{\text{cf}} $ is the counterfactual dataset and $ \delta $ controls perturbation magnitude. Perturbations, such as modifying $ P_t $ with noise $ \epsilon_t \sim \mathcal{N}(0, \sigma^2) $ or adjusting $ F_t $ and $ N_t $, preserve statistical relationships. Strategy performance is assessed as:
\begin{equation}
R = \phi(C, D), \quad R_{\text{cf}} = \phi(C, D_{\text{cf}}),
\end{equation}
where $ \phi $ evaluates strategy $ C $ on datasets $ D $ and $ D_{\text{cf}} $. Information leakage is quantified using Prediction Consistency (PC), Confidence Invariance (CI), and Input Dependency Score (IDS). Strategies with high PC and CI but low IDS rely on memorized patterns. CS optimizes strategies to minimize leakage:
\begin{equation}
C^* = \arg\min_{C} \{ \alpha \cdot \text{PC}(C) + \beta \cdot \text{CI}(C) - \gamma \cdot \text{IDS}(C) \},
\end{equation}
where $ \alpha $, $ \beta $, and $ \gamma $ are weights, ensuring $ C^* $ is driven by real-time inputs and robust against leakage.


\begin{algorithm}[t]
\caption{Workflow of Our FactFin}
\label{alg:FactFin}
\setstretch{1.15}
\begin{algorithmic}[1]
\Require Market state $S_t = \{P_t, F_t, N_t\}$, dataset $D$
\State $S_t' \gets \text{RAG}(S_t)$ \Comment{Factor extraction}
\State $C \gets \text{SCG}(S_t', P)$ \Comment{Generate initial strategy}
\While{not converged}
    \State $C \gets \text{MCTS}(C, D)$ \Comment{Optimize strategy}
    \State $D_{\text{cf}} \gets \text{Perturb}(D, \delta)$ \Comment{Counterfactual data}
    \State Evaluate $\text{PC}, \text{CI}, \text{IDS}$ on $\phi(C, D)$ and $\phi(C, D_{\text{cf}})$
    \State Update $C \gets \arg\min \left\{
        \begin{aligned}
        &\alpha \cdot \text{PC}(C) + \beta \cdot \text{CI}(C) \\
        &\quad - \gamma \cdot \text{IDS}(C)
        \end{aligned}
    \right\}$
\EndWhile
\Ensure Optimized strategy $C^*$
\end{algorithmic}
\end{algorithm}

\section{Experimental Results}

\subsection{Seetings}

\subsubsection{Dataset}
\label{subsec:dataset}

We evaluate our FactFin framework via six financial assets: U.S. equities (AAPL, NVDA, TSLA), Chinese equity (BYD, 002594.SZ), Hong Kong equity (Tencent, 0700.HK), and cryptocurrency (Bitcoin). Sourced from Yahoo Finance and Alpaca News API, the dataset spans January 1, 2020, to June 30, 2025, including price data with volume and turnover, news, and counterfactual scenarios to assess information leakage. 


\subsubsection{Evaluation Metrics}
\label{subsec:metrics}

The profitability and risk are evaluated using Total Return (TR), Sharpe Ratio (SR), and downside risk using Maximum Drawdown (MDD), defined in Section~\ref{subsec:temp_seg}. Information leakage is evaluated through Prediction Consistency (PC), Confidence Invariance (CI), and Input Dependency Score (IDS),  defined in Section~\ref{subsec:counterfactual}, which measure stability, sensitivity, and data dependence during counterfactual changes.




\subsubsection{Baselines}
\label{subsec:baselines}
1) Financial models: FinGPT~\cite{yang2023fingpt}, Fin-LLaMA~\cite{Fin-LLAMA}, InvestLM~\cite{yang2023investlm}; 2) Single-agent systems: FinMem~\cite{yu2023finmem}, FinAgent~\cite{zhang2024multimodal}, QuantAgent~\cite{wang2024quantagentseekingholygrail}; 3) Multi-agent systems: TradingAgents~\cite{xiao2025tradingagentsmultiagentsllmfinancial}, \\HedgeAgents~\cite{10.1145/3701716.3715232}, FinRobot~\cite{zhou2024finrobot}). All baselines are implemented per their original specifications\footnote{More details are provided in Appendix.}.

\subsubsection{Implementation Details}
\label{subsec:impl_details}

The action space \( A_t \) includes three discrete trading actions for individual stocks: buy, sell, and hold, executed with standard transaction costs and realistic slippage models. FactFin and all LLM-based agents in baselines use GPT-4o as the backbone with a temperature of 0.7 for balanced consistency and creativity. RAG adopts the text-embedding-3-large model\cite{openai2023textembedding3} with top-k=5, while MCTS is configured with a search depth of 10 and UCB exploration parameter \( c=0.5 \). 

\begin{table*}[t]
\centering
\caption{Performance comparison of FactFin and baselines across six assets (July 2024 to June 2025). \textbf{\textcolor{red}{Bold}} represents optimal performance, while \underline{\textbf{\textcolor{blue}{Bold}}} represents suboptimal.}
\label{tab:overall_performance}
\begin{adjustbox}{width=0.95\textwidth} 
\small 
\setlength{\tabcolsep}{2pt} 
\begin{tabular}{clcccccccccccccccccccc}
\toprule
& & \multicolumn{3}{c}{\textbf{AAPL}} & \multicolumn{3}{c}{\textbf{NVDA}} & \multicolumn{3}{c}{\textbf{TSLA}} & \multicolumn{3}{c}{\textbf{BYD}} & \multicolumn{3}{c}{\textbf{Tencent}} & \multicolumn{3}{c}{\textbf{Bitcoin}} \\
\cmidrule(lr){3-5} \cmidrule(lr){6-8} \cmidrule(lr){9-11} \cmidrule(lr){12-14} \cmidrule(lr){15-17} \cmidrule(lr){18-20}
\multirow{-2}{*}{\textbf{Categories}} & \multirow{-2}{*}{\textbf{Models}} & TR $\uparrow$ & SR $\uparrow$ & MDD $\downarrow$ & TR $\uparrow$ & SR $\uparrow$ & MDD $\downarrow$ & TR $\uparrow$ & SR $\uparrow$ & MDD $\downarrow$ & TR $\uparrow$ & SR $\uparrow$ & MDD $\downarrow$ & TR $\uparrow$ & SR $\uparrow$ & MDD $\downarrow$ & TR $\uparrow$ & SR $\uparrow$ & MDD $\downarrow$ \\
\midrule
Market & B\&H & -3.36\% & 0.05 & 33.43\% & 27.81\% & 0.72 & 36.89\% & 57.88\% & 0.98 & 53.77\% & 32.43\% & 0.96 & 19.56\% & 36.82\% & 1.12 & 23.49\% & 70.40\% & 1.10 & 28.11\% \\
\midrule
& FinGPT & 14.48\% & 0.59 & 29.05\% & 28.01\% & 0.73 & 32.62\% & 42.58\% & 0.85 & 52.37\% & 10.21\% & 0.46 & 21.31\% & -5.93\% & -0.05 & 23.28\% & 42.81\% & 0.85 & 24.61\% \\
& Fin-LLaMA & -20.05\% & -0.63 & 37.76\% & 16.27\% & 0.55 & 32.65\% & 63.15\% & 1.06 & 53.71\% & -9.61\% & -0.12 & 22.23\% & 23.64\% & 0.82 & 23.47\% & 21.04\% & 0.53 & 31.19\% \\
\multirow{-3}{*}{Fin-LLM} & InvestLM & -9.25\% & -0.21 & 32.86\% & 38.18\% & 0.89 & 28.40\% & 50.15\% & 0.96 & \underline{\textbf{\textcolor{blue}{36.88\%}}} & -4.11\% & 0.03 & 23.17\% & -4.85\% & -0.06 & 19.92\% & 54.88\% & 1.01 & 26.66\% \\
\midrule
& FinMem & -5.68\% & -0.03 & 32.11\% & 35.88\% & 0.84 & 33.42\% & 52.32\% & 0.95 & 44.63\% & 26.83\% & 0.88 & 19.55\% & 27.77\% & 0.97 & 17.40\% & 72.68\% & 1.26 & \underline{\textbf{\textcolor{blue}{19.68\%}}} \\
& FinAgent & \underline{\textbf{\textcolor{blue}{27.79\%}}} & \underline{\textbf{\textcolor{blue}{0.99}}} & 21.52\% & 54.86\% & 1.09 & 27.78\% & 79.01\% & 1.24 & -9.36\% & 32.91\% & 1.04 & 20.17\% & 44.18\% & 1.29 & 23.44\% & 94.63\% & 1.38 & 24.53\% \\
\multirow{-3}{*}{Single-Agent} & QuantAgent & 6.60\% & 0.36 & 29.39\% & 49.83\% & 1.03 & 30.12\% & 74.26\% & 1.21 & 42.35\% & 31.74\% & 1.03 & \underline{\textbf{\textcolor{blue}{18.01\%}}} & 31.15\% & 1.08 & 19.95\% & 90.56\% & 1.35 & 32.98\% \\
\midrule
& TradingAgents & 12.89\% & 0.55 & 27.26\% & \underline{\textbf{\textcolor{blue}{59.28\%}}} & \underline{\textbf{\textcolor{blue}{1.17}}} & \underline{\textbf{\textcolor{blue}{25.99\%}}} & 106.01\% & 1.47 & 36.90\% & 43.33\% & 1.26 & 19.53\% & 42.77\% & 1.38 & 24.18\% & 88.39\% & 1.37 & 32.96\% \\
& HedgeAgents & 16.06\% & 0.68 & \underline{\textbf{\textcolor{blue}{12.12\%}}} & 54.09\% & 1.10 & 28.39\% & \underline{\textbf{\textcolor{blue}{115.09\%}}} & \underline{\textbf{\textcolor{blue}{1.48}}} & 48.79\% & \underline{\textbf{\textcolor{blue}{57.96\%}}} & \underline{\textbf{\textcolor{blue}{1.49}}} & 18.72\% & \underline{\textbf{\textcolor{blue}{66.31\%}}} & \underline{\textbf{\textcolor{blue}{1.85}}} & 17.67\% & \underline{\textbf{\textcolor{blue}{134.36\%}}} & \underline{\textbf{\textcolor{blue}{1.71}}} & 22.12\% \\
\multirow{-3}{*}{Multi-Agents} & FinRobot & 21.76\% & 0.81 & 21.54\% & 43.98\% & 0.72 & 36.65\% & 96.32\% & 1.34 & 49.99\% & 38.49\% & 1.09 & 20.53\% & 57.54\% & 1.81 & \underline{\textbf{\textcolor{blue}{14.56\%}}} & 109.18\% & 1.51 & 21.67\% \\
\midrule
\rowcolor[gray]{0.9}\textbf{Ours} & \textbf{FactFin} & \textbf{\textcolor{red}{36.70\%}} & \textbf{\textcolor{red}{1.22}} & \textbf{\textcolor{red}{11.57\%}} & \textbf{\textcolor{red}{71.34\%}} & \textbf{\textcolor{red}{1.29}} & \textbf{\textcolor{red}{24.25\%}} & \textbf{\textcolor{red}{165.01\%}} & \textbf{\textcolor{red}{1.83}} & \textbf{\textcolor{red}{31.54\%}} & \textbf{\textcolor{red}{84.24\%}} & \textbf{\textcolor{red}{2.09}} & \textbf{\textcolor{red}{16.01\%}} & \textbf{\textcolor{red}{81.37\%}} & \textbf{\textcolor{red}{2.31}} & \textbf{\textcolor{red}{14.14\%}} & \textbf{\textcolor{red}{171.46\%}} & \textbf{\textcolor{red}{2.03}} & \textbf{\textcolor{red}{16.59\%}} \\
\midrule
\rowcolor[gray]{0.9}\multicolumn{2}{c}{\textbf{Improvement(\%)}} & 32.06\% & 23.23\% & 4.54\% & 20.34\% & 10.26\% & 6.69\% & 43.37\% & 19.13\% & 14.48\% & 45.34\% & 40.27\% & 11.10\% & 22.71\% & 24.86\% & 2.88\% & 27.61\% & 18.71\% & 15.70\% \\
\bottomrule
\end{tabular}%
\end{adjustbox}
\end{table*}

\begin{figure*}[t]
\centering
\includegraphics[width=0.96\textwidth]{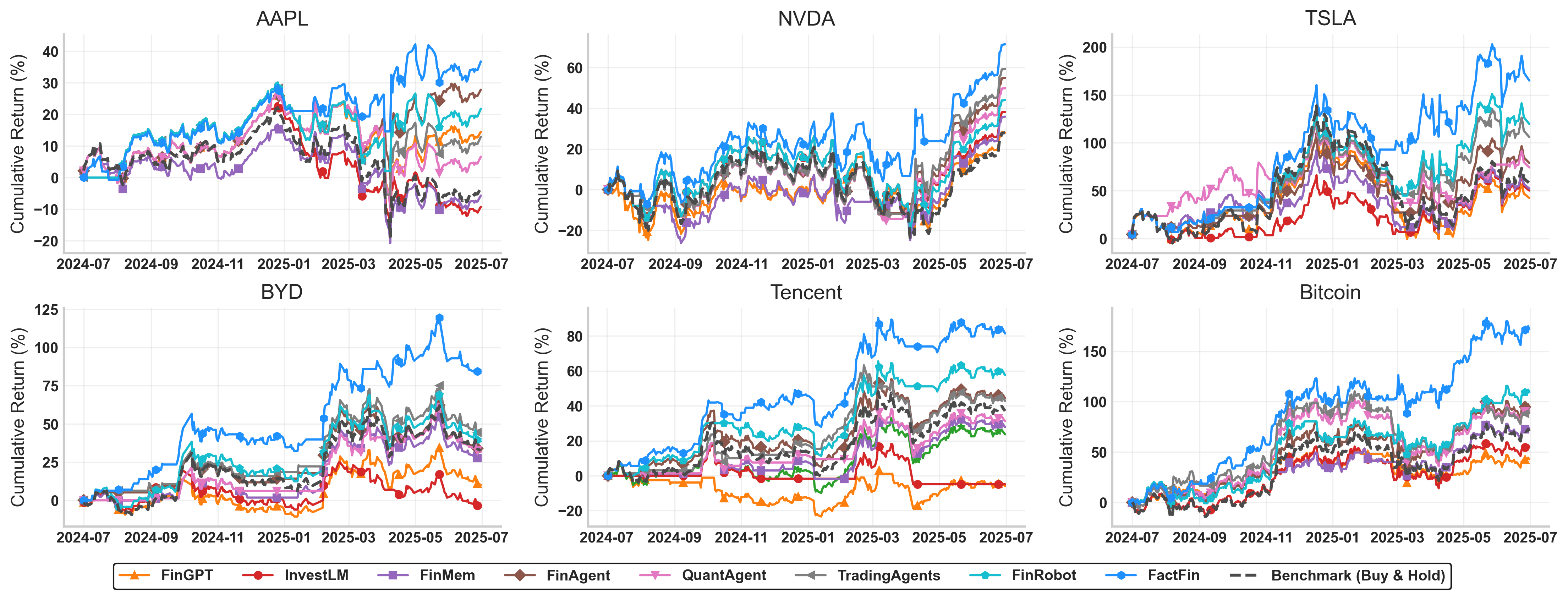}
\caption{Performance comparison over time between FactFin and other benchmarks across all assets.}
\label{fig:overall_cr}
\vspace{-0.2cm}
\end{figure*}

\subsection{Overall Performance}
\label{subsec:overall_performance}

Table~\ref{tab:overall_performance} evaluates FactFin against nine baseline methods across six assets from July 1, 2024, to June 30, 2025, after the release of GPT-4o. 

The following observations are made: 
1) Financial models exhibit highly unstable performance across markets; for instance, FinGPT achieves an excess return of 17.84\% for AAPL but -42.75\% for Tencent. This inconsistency arises from training data biases favoring familiar assets, resulting in poor generalization.
2) Single-agent systems demonstrate improved returns and risk management with more stable performance across assets; FinAgent achieves a TR of 27.79\% and SR of 0.99 for AAPL, the best among nine baselines. This improvement is due to tool invocation for market analysis, outperforming prompt-dependent Fin-LLM.
3) Multi-agent systems enhance performance through collaborative task decomposition; TradingAgents achieves a TR of 59.28\%, SR of 1.17, and MDD of 25.99\% for NVDA, the best among nine baselines, but FinRobot suffers from a high MDD of-49.99\% for TSLA. This is because dynamic weight allocation adapts to market changes, yet over-analysis of non-informative data, such as macro news.
4) Our FactFin consistently outperforms all baselines across all assets, achieving an average improvement of 31.91\% in TR, 22.74\% in SR, and 9.23\% in MDD, with robust cumulative returns, as shown in Figure~\ref{fig:overall_cr}. This superiority comes from a comprehensive market analysis using factorization and data-driven strategies, combined with counterfactual reasoning and strategy evolution, thus outperforming baselines.

\subsection{Information Leakage Mitigation}
\label{subsec:leakage_mitigation}

To evaluate the ability of FactFin and baselines to mitigate information leakage, we conducted experiments prior to the LLM training cutoff, constructing 50 counterfactual scenarios for each asset. Table~\ref{tab:leakage_metrics} presents the information leakage metrics across all models and assets. 

The following observations can be made: 1) Financial language models exhibit severe information leakage, with InvestLM showing high average PC 0.8789 and average CI 0.8964 and low average IDS 0.1912 across six assets. This is due to fine-tuning on financial datasets, which leads to memorization of specific events rather than learning generalizable patterns, resulting in poor performance in counterfactual scenarios. 2) Single-agent systems show reduced information leakage, as demonstrated by QuantAgent achieving optimal PC 0.6775 and IDS 0.3868 among nine baselines on BYD, leveraging tool invocation to mitigate memory reliance, though long contexts lead to moderate leakage. 3) Multi-agent systems further reduce information leakage, benefiting from collaborative task decomposition and information cross-validation, but residual leakage persists. 4) Our FactFin, through Monte Carlo Tree Search and a Counterfactual Simulator, uses the LLM as a strategy generator rather than a direct decision-maker, substantially reducing PC and CI while increasing IDS, achieving the lowest leakage across all assets. 
These results show that FactFin reliably predicts across various market conditions 

\begin{table*}[t]
\centering
\caption{Information leakage metrics across six assets.\textbf{Bold} indicates the best result; \underline{Underline} indicates the second-best.}
\label{tab:leakage_metrics}
\begin{adjustbox}{width=0.98\textwidth} 
\small 
\setlength{\tabcolsep}{2pt} 
\begin{tabular}{clccccccccccccccccccc}
\toprule
& & \multicolumn{3}{c}{\textbf{AAPL}} & \multicolumn{3}{c}{\textbf{NVDA}} & \multicolumn{3}{c}{\textbf{TSLA}} & \multicolumn{3}{c}{\textbf{BYD}} & \multicolumn{3}{c}{\textbf{Tencent}} & \multicolumn{3}{c}{\textbf{Bitcoin}} \\
\cmidrule(lr){3-5} \cmidrule(lr){6-8} \cmidrule(lr){9-11} \cmidrule(lr){12-14} \cmidrule(lr){15-17} \cmidrule(lr){18-20}
\multirow{-2}{*}{\textbf{Categories}} & \multirow{-2}{*}{\textbf{Models}} & PC $\downarrow$ & CI $\downarrow$ & IDS $\uparrow$ & PC $\downarrow$ & CI $\downarrow$ & IDS $\uparrow$ & PC $\downarrow$ & CI $\downarrow$ & IDS $\uparrow$ & PC $\downarrow$ & CI $\downarrow$ & IDS $\uparrow$ & PC $\downarrow$ & CI $\downarrow$ & IDS $\uparrow$ & PC $\downarrow$ & CI $\downarrow$ & IDS $\uparrow$ \\
\midrule
Market & FinGPT & 0.8239 & 0.9011 & 0.1851 & 0.9172 & 0.9240 & 0.1233 & 0.7587 & 0.8189 & 0.2818 & 0.8091 & 0.8393 & 0.3014 & 0.7188 & 0.7596 & 0.3316 & 0.6955 & 0.7277 & 0.3513 \\
\midrule
\multirow{3}{*}{Fin-LLM}
& Fin-LLaMA & 0.8785 & 0.9288 & 0.2127 & 0.8835 & 0.9113 & 0.2139 & 0.8355 & 0.8681 & 0.2425 & 0.7802 & 0.8123 & 0.3273 & 0.8425 & 0.8701 & 0.2259 & 0.7872 & 0.8129 & 0.2618 \\
& InvestLM & 0.9187 & 0.9223 & 0.1618 & 0.9213 & 0.9322 & 0.1716 & 0.8527 & 0.8819 & 0.2123 & 0.8285 & 0.8391 & 0.2517 & 0.8612 & 0.8908 & 0.1959 & 0.8911 & 0.9121 & 0.1537 \\
\midrule
\multirow{4}{*}{LLM-Agent}
& FinMem & 0.8578 & 0.8662 & 0.2531 & 0.8123 & 0.8235 & 0.2809 & 0.7651 & 0.8032 & 0.3014 & 0.7912 & 0.8073 & 0.2833 & 0.7716 & 0.8125 & 0.3268 & 0.8235 & 0.8481 & 0.2918 \\
& FinAgent & 0.7252 & 0.7316 & 0.3402 & 0.7566 & 0.7907 & 0.3825 & 0.7408 & 0.7524 & 0.3635 & 0.7395 & 0.7735 & 0.3341 & 0.7219 & 0.7395 & 0.3643 & 0.7574 & 0.7802 & 0.3639 \\
& QuantAgent & 0.7436 & 0.7768 & 0.3251 & 0.7723 & 0.8168 & 0.3526 & 0.7262 & 0.7659 & 0.3422 & \underline{0.6775} & 0.7079 & \underline{0.3868} & 0.7358 & 0.7762 & 0.3425 & 0.7976 & 0.8038 & 0.3276 \\
\midrule
\multirow{3}{*}{LLM-MultiAgents}
& TradingAgents & 0.6882 & 0.7252 & \underline{0.4248} & \underline{0.6671} & \underline{0.6975} & 0.4413 & 0.6816 & 0.7003 & \underline{0.4151} & 0.6912 & 0.7306 & 0.3647 & 0.6728 & 0.7031 & \underline{0.3849} & 0.7095 & 0.7334 & 0.3945 \\
& HedgeAgents & \underline{0.6594} & \underline{0.6942} & 0.4052 & 0.6808 & 0.7012 & \underline{0.4688} & 0.7163 & 0.7367 & 0.3855 & 0.6786 & \underline{0.6927} & 0.3851 & \underline{0.6443} & \underline{0.6728} & 0.3653 & \underline{0.6755} & \underline{0.6971} & \underline{0.4229} \\
& FinRobot & 0.7267 & 0.7591 & 0.3845 & 0.7321 & 0.7414 & 0.3946 & \underline{0.6591} & \underline{0.6629} & 0.3773 & 0.6892 & 0.7248 & 0.3549 & 0.7166 & 0.7325 & 0.3561 & 0.7269 & 0.7343 & 0.3817 \\
\midrule
\rowcolor[gray]{0.9}Ours & \textbf{FactFin} & \textbf{0.3115} & \textbf{0.2548} & \textbf{0.7781} & \textbf{0.2842} & \textbf{0.2645} & \textbf{0.7613} & \textbf{0.3427} & \textbf{0.3057} & \textbf{0.7544} & \textbf{0.2424} & \textbf{0.2273} & \textbf{0.8279} & \textbf{0.2612} & \textbf{0.2509} & \textbf{0.7726} & \textbf{0.2843} & \textbf{0.3146} & \textbf{0.7847} \\
\midrule
\rowcolor[gray]{0.9}\multicolumn{2}{c}{\textbf{Improvement(\%)}} & 52.77\% & 63.30\% & 83.17\% & 57.39\% & 62.08\% & 62.39\% & 48.00\% & 53.89\% & 81.74\% & 64.22\% & 67.18\% & 114.04\% & 59.45\% & 62.70\% & 100.73\% & 57.91\% & 54.87\% & 85.55\% \\
\bottomrule
\end{tabular}%

\end{adjustbox}
\end{table*}

\begin{table*}[t]
\centering
\caption{Ablation studies over different components. $\checkmark$ indicates the component is added to FactFin.}
\label{tab:ablation}

\setlength{\tabcolsep}{2pt}

\small
\resizebox{\textwidth}{!}{%
\begin{tabular}{cccccccccccccccc}
\toprule
\multicolumn{4}{c}{\textbf{Components}} & \multicolumn{6}{c}{\textbf{AAPL}} & \multicolumn{6}{c}{\textbf{TSLA}} \\
\cmidrule(lr){1-4} \cmidrule(lr){5-10} \cmidrule(lr){11-16}
CS & MCTS & RAG & SCG & TR $\uparrow$ & SR $\uparrow$ & MDD $\downarrow$ & PC $\downarrow$ & CI $\downarrow$ & IDS $\uparrow$ & TR $\uparrow$ & SR $\uparrow$ & MDD $\downarrow$ & PC $\downarrow$ & CI $\downarrow$ & IDS $\uparrow$ \\
\midrule
 &  &  & \checkmark & 8.77\% & 0.43 & 24.34\% & 0.6213 & 0.6457 & 0.4361 & 78.42\% & 1.22 & 40.42\% & 0.6703 & 0.6319 & 0.3857 \\
 &  & \checkmark & \checkmark & 13.36\% & 0.56 & 23.65\% & 0.5529 & 0.5201 & 0.4903 & 104.38\% & 1.44 & 37.45\% & 0.5852 & 0.5638 & 0.4293 \\
 & \checkmark & \checkmark & \checkmark & 28.12\% & 0.99 & 16.20\% & 0.4858 & 0.5026 & 0.5348 & 130.93\% & 1.64 & 33.52\% & 0.4927 & 0.4481 & 0.5299 \\
\textbf{\checkmark} & \textbf{\checkmark} & \textbf{\checkmark} & \textbf{\checkmark} & \textbf{36.70\%} & \textbf{1.22} & \textbf{11.57\%} & \textbf{0.3115} & \textbf{0.2548} & \textbf{0.7781} & \textbf{165.01\%} & \textbf{1.83} & \textbf{31.54\%} & \textbf{0.3427} & \textbf{0.3057} & \textbf{0.7544} \\
\bottomrule
\end{tabular}%
}
\end{table*}

\subsection{Ablation Studies}
\label{subsec:ablation_studies}

In Table~\ref{tab:ablation}, we study the effectiveness of the Strategy Code Generator (SCG), Retrieval-Augmented Generation (RAG), Monte Carlo Tree Search (MCTS), and Counterfactual Simulator (CS) in FactFin.
1)When using only SCG, FactFin exhibits low financial performance and high information leakage. Although SCG outperforms direct LLM decision-making, code generation alone is insufficient to effectively mitigate leakage. 2)Adding RAG significantly improves TR and SR for both assets and reduces leakage, as structured market factor extraction enhances information utilization. 
3) Introducing MCTS further increases TR from 13.36\% to 28.12\% for AAPL and from 104.38\% to 130.93\% for TSLA, while reducing MDD from 23.65\% to 16.20\% for AAPL and from 37.45\% to 33.52\% for TSLA, substantially contribute to improved returns and risk reduction, with leakage moderately alleviated, though still at a high level.
4)The full model, including CS, achieves optimal performance across all metrics, with information leakage most effectively mitigated. The CS component is crucial for detecting and correcting leaks, allowing FactFin to avoid memorized patterns, ensuring strong performance and minimal data leakage.

\begin{table}[t]
\centering
\caption{Performance comparison of LLM backbones.}
\label{tab:llm_comparison}

\setlength{\tabcolsep}{3pt}

\small 
\resizebox{\columnwidth}{!}{%
\begin{tabular}{lcccccc}
\toprule
\textbf{Models} & \textbf{TR} $\uparrow$ & \textbf{SR} $\uparrow$ & \textbf{MDD} $\downarrow$ & \textbf{PC} $\downarrow$ & \textbf{CI} $\downarrow$ & \textbf{IDS} $\uparrow$ \\
\midrule
Qwen2.5-72B-Instruct & 93.12\% & 1.65 & 20.34\% & 0.3098 & 0.2897 & 0.7623 \\
LLaMA-3.1 405B & 91.56\% & 1.62 & 20.89\% & 0.3156 & 0.2956 & 0.7567 \\
DeepSeek-V3 & 99.78\% & 1.76 & \textbf{17.89\%} & 0.2823 & \textbf{0.2654} & 0.7692 \\
Claude-Sonnet-3.5 & 98.23\% & 1.74 & 18.45\% & \textbf{0.2756} & 0.2689 & 0.7633 \\
Gemini-2.0-Flash & 89.45\% & 1.58 & 21.78\% & 0.3212 & 0.3045 & 0.7456 \\
GPT-4o & \textbf{101.69\%} & \textbf{1.79} & 19.02\% & 0.2877 & 0.2696 & \textbf{0.7798} \\
\bottomrule
\end{tabular}%
}
\end{table}

\subsection{LLM Backbone Comparison}
\label{subsec:llm_comparison}

To assess FactFin's adaptability, we tested six state-of-the-art LLMs as backbones, 
Table~\ref{tab:llm_comparison} presents the performance. Closed-source models outperformed open-source ones in financial metrics and leakage control
GPT-4o led in TR (101.69\%), SR (1.79), and IDS (0.7798), while Claude-Sonnet-3.5 (PC: 0.2756) and DeepSeek-V3 (CI: 0.2654) showed best leakage resistance. All models showed strong financial returns and minimal leakage, confirming FactFin's robustness across architectures.
\section{Related Works}
\label{subsec:apprew} 

\textbf{LLM-based financial system}
Quantitative finance is an interdisciplinary field that integrates finance with mathematical and statistical methods to address complex financial challenges \cite{kou2019machine,kanamura2021pricing}. With the advent of LLMs, an increasing number of researchers are leveraging cutting-edge technologies in finance. \citet{yang2023fingpt} proposed FinGPT, which enables a thorough understanding of financial events and facilitates news analysis. \citet{li2024finreport} introduced FinRport, a framework that amalgamates diverse information to generate financial reports on a regular basis. Compared to conventional models \cite{YU2011367,wang2021deeptrader}, these LLM-based approaches improve the accuracy and efficiency of market forecasting. However, these methods have yet to be fully learn from real-world fund companies, and essential components have not been included.

\noindent\textbf{Multi-Agent Framewrok}
LLM-based agent systems, leveraging their cognitive and generative capabilities, have the ability to perform a range of complex tasks, including knowledge integration, information retention, logical reasoning, and strategic planning \cite{sumers2023cognitive,pan2023llms}. Furthermore, initiatives based on multi-agent systems, such as ``The Sims'' from Stanford University \cite{park2023generative}, have demonstrated the formidable power of collective intelligence. Through the collaboration of multiple agents, multi-agent systems are expected to make significant contributions in fields such as finance \cite{zhang2024multimodal}, offering innovative approaches and sophisticated solutions for complex challenges \cite{hong2023metagpt,wu2023bloomberggpt}.

\section{Conclusion}

In this paper, we systematically investigated and carefully addressed the critical and often-overlooked ``profit mirage'' phenomenon in LLM-based financial agents. Our main contributions are twofold: First, we developed FinLake-Bench, a comprehensive and rigorous benchmark that rigorously evaluates information leakage across multiple dimensions. Second, we proposed FactFin, a novel framework that leverages counterfactual strategy to enhance the robustness. Through extensive empirical validation, we convincingly demonstrated that our FactFin significantly outperforms existing approaches in out-of-sample scenarios, achieving superior risk-adjusted returns while effectively mitigating the persistent information leakage problem. 

\newpage
\clearpage
\bibliographystyle{ACM-Reference-Format}
\bibliography{fineleak_www}

\appendix

\definecolor{promptBlueBack}{HTML}{EBF5FF}
\definecolor{promptBlueFrame}{HTML}{003366}
\definecolor{promptGreenBack}{HTML}{F0FFF0}
\definecolor{promptGreenFrame}{HTML}{004D00}
\definecolor{promptOrangeBack}{HTML}{FFF5E6}
\definecolor{promptOrangeFrame}{HTML}{D35400} 

\newtcolorbox{systempromptbox}[1]{
    colback=promptBlueBack,   
    colframe=promptBlueFrame, 
    title=#1,                 
    fonttitle=\bfseries, 
    boxrule=0.5mm, 
    arc=2mm, 
    left=2mm,
    right=2mm, 
    top=2mm,
    bottom=2mm 
}
\newtcolorbox{userpromptbox}[1]{
    colback=promptGreenBack,    
    colframe=promptGreenFrame,  
    title=#1,             
    fonttitle=\bfseries, 
    boxrule=0.5mm, 
    arc=2mm, 
    left=2mm,
    right=2mm, 
    top=2mm,
    bottom=2mm 
}

\newtcolorbox{examplebox}[1]{
    colback=promptOrangeBack,    
    colframe=promptOrangeFrame,  
    title=#1,             
    fonttitle=\bfseries, 
    boxrule=0.5mm, 
    arc=2mm, 
    left=2mm,
    right=2mm, 
    top=2mm,
    bottom=2mm 
}







\section{Dataset}
~\label{subsec:appdataset}

We evaluate our FinLeak framework via six financial assets: U.S. equities (AAPL, NVDA, TSLA), Chinese equity (BYD, 002594.SZ), Hong Kong equity (Tencent, 0700.HK), and cryptocurrency (Bitcoin). Sourced from Yahoo Finance and Alpaca News API, the dataset spans January 1, 2020, to June 30, 2025, including price data with volume and turnover, news, and counterfactual scenarios to assess information leakage. Table~\ref{tab:dataset} is the key statistics.

\begin{table}[h]
\centering
\scriptsize
\setlength{\tabcolsep}{2pt}
\renewcommand{\arraystretch}{0.9}
\caption{Dataset statistics detailing the chronological period and the number of each data source for each asset}
\label{tab:dataset}
\begin{tabular}{lcccccc}
\toprule
\textbf{Metric} & \textbf{AAPL} & \textbf{NVDA} & \textbf{TSLA} & \textbf{BYD} & \textbf{Tencent} & \textbf{Bitcoin} \\
\midrule
Trading Date & \multicolumn{6}{c}{Jan 1, 2020 -- Jun 30, 2025} \\
& \multicolumn{6}{c}{(1380/1380/1380/1329/1350/2008 days)} \\
Asset Price & \multicolumn{6}{c}{days $\times$ (open, high, low, close,} \\
& \multicolumn{6}{c}{adj\_close, vol, turn)} \\
Asset News & 26551 & 28432 & 31677 & 17423 & 18816 & 20213 \\
Counterfactual Scenarios & 253 & 267 & 271 & 229 & 243 & 279 \\
\bottomrule
\end{tabular}
\end{table}

\begin{table*}[ht]
\centering
\small
\caption{Detailed FinLeak-Bench Case Study: Model Responses and Scoring}
\label{tab:finleak-casestudy}
\begin{tabular}{p{2.2cm} p{9cm} p{2cm}}
\toprule
\multicolumn{3}{l}{\textbf{1. Price Inquiry}} \\
\midrule
\textbf{Question} & What was Microsoft's highest price on November 8, 2022? & \\
\textbf{Standard Answer} & \$243.74 & \\
\textbf{Scoring Rule} & 1 point if answer $\in$ [241.30, 246.18]; 0.5 point if $\in$ [236.43, 251.05]; otherwise 0 & \\
\midrule
\textbf{Model} & \textbf{Response} & \textbf{Score} \\
GPT-4o & Microsoft's intraday high reached \$243.90 on November 8, 2022. & 1 point \\
Claude-Sonnet-3.7 & The highest price for Microsoft was \$242.50 on November 8, 2022. & 1 point \\
Grok-3 & Microsoft's peak price on that day was \$244.00. & 1 point \\
\midrule
\multicolumn{3}{l}{\textbf{2. Event Impact}} \\
\midrule
\textbf{Question} & How did China’s COVID policy adjustment on December 7, 2022, affect Chinese stocks? & \\
\textbf{Standard Answer} & Positive impact: Market optimism drove over 40\% rebound since October. & \\
\textbf{Scoring Rule} & 1 point for mentioning policy easing \& >40\% rally; 0.5 point if only general positive impact mentioned & \\
\midrule
\textbf{Model} & \textbf{Response} & \textbf{Score} \\
GPT-4o & The policy easing triggered strong investor optimism, leading to a sharp rebound exceeding 40\% in Chinese stocks since October. & 1 point \\
Claude-Sonnet-3.7 & China's easing of COVID restrictions on December 7 spurred a market rebound of over 40\% for Chinese equities. & 1 point \\
Grok-3 & The adjustment in COVID policies was received positively, with investor confidence pushing Chinese stocks higher. & 0.5 point \\
\midrule
\multicolumn{3}{l}{\textbf{3. Trend Prediction}} \\
\midrule
\textbf{Question} & How did Meta perform over 6 weeks following September 20, 2022? & \\
\textbf{Standard Answer} & Slight rise then persistent decline, cumulative drop of 23.70\%. & \\
\textbf{Scoring Rule} & 0.5 point for correct overall downward trend; +0.5 for magnitude prediction within $\pm$5\% error & \\
\midrule
\textbf{Model} & \textbf{Response} & \textbf{Score} \\
GPT-4o & Meta's stock had a brief initial increase but then fell steadily, losing approximately 24\% over six weeks. & 1 point \\
Claude-Sonnet-3.7 & Meta experienced a slight rise before dropping around 22\% in total over the following six weeks. & 1 point \\
Grok-3 & After a minor early uptick, Meta's shares declined persistently, ending with a cumulative loss close to 23.5\%. & 1 point \\
\midrule
\multicolumn{3}{l}{\textbf{4. Market Performance}} \\
\midrule
\textbf{Question} & Which stock had the highest return in the S\&P 500 on June 16, 2022? & \\
\textbf{Standard Answer} & NEM (Newmont Corporation) was the best performer. & \\
\textbf{Scoring Rule} & 1 point for NEM, KR, EPAM, LH, WMT (within ±1\% error); 0.5 point for ABBV, MKTX, PG, UDR, EW, EQR, NWSA, INCY, CF, EXR (within ±3\% error) & \\
\midrule
\textbf{Model} & \textbf{Response} & \textbf{Score} \\
GPT-4o & WMT led the S\&P 500 on June 16, 2022, posting the highest return of the day. & 1 point \\
Claude-Sonnet-3.7 & NEM was among the top performers in the S\&P 500 on that date, registering the largest gain. & 1 point \\
Grok-3 & NWSA was one of the top gainers in the S\&P 500 on June 16, contributing significantly to the day's rally. & 0.5 point \\
\bottomrule
\end{tabular}
\end{table*}

\section{FinLeak-Bench Case Study}
\label{subsec:appfb} 
FinLeak-Bench represents a comprehensive evaluation benchmark designed to systematically assess information leakage and memorization patterns in large language models when applied to financial prediction tasks. The benchmark comprises 2,000 carefully curated financial question-answer pairs spanning from January 2022 to June 2023, a critical period that overlaps with the training cutoff dates of most contemporary LLMs. This temporal alignment enables precise detection of memorization-based responses versus genuine predictive capabilities.

The construction of FinLeak-Bench follows a rigorous methodology designed to capture different dimensions of financial knowledge that LLMs might have memorized during training. We systematically categorize financial information into four distinct types that represent different levels of memorization complexity: specific data point recall (price inquiries), temporal pattern recognition (trend predictions), causal relationship memory (event impacts), and comparative ranking recall (market performance). Each category is designed to probe specific aspects of how LLMs might rely on memorized training data rather than input-driven reasoning.
We provide a concise case study from FinLeak-Bench to illustrate the evaluation and scoring procedure (Table~\ref{tab:finleak-casestudy}).

\begin{table*}[t]
\centering
\caption{Integrated Key Perturbations Across Counterfactual Scenarios. Unchanged fields omitted for brevity.}
\label{tab:integrated}
\small
\renewcommand{\arraystretch}{1.2} 
\setlength{\tabcolsep}{8pt}       
\begin{tabularx}{\textwidth}{>{\raggedright\arraybackslash}p{0.22\textwidth} p{0.1\textwidth} X X}
\toprule
\textbf{Case Study} & \textbf{Element} & \textbf{Original} & \textbf{Counterfactual} \\
\midrule
\ NVDA Earnings (May 25, 2022)
& market\_news 
& Revenue: \$8.29B (+46\% YoY); Data Center: \$3.75B (+83\% YoY); Gaming: \$3.62B (+31\% YoY); Q2 outlook: \$8.10B; Strong data center growth 
& Revenue: \$7.64B (below expectations); Data Center: \$3.05B (below expectations); Gaming: \$2.95B (supply chain issues); Q2 outlook: \$7.50B (cautious); Concerns over slowing growth \\
\midrule
\ TSLA Trend Reversal (Oct 19, 2022) 
& price\_data (5-day) 
& [221.72, 204.99, 219.35, 220.19, 222.04] (upward trend) 
& [239.09, 232.83, 226.11, 229.56, 226.62] (downward trend) \\
& technical\_indicators 
& RSI: 28.0 (Oversold); MACD: -18.26 (Bearish); 50-day MA: 274.17; 200-day MA: 283.23 
& RSI: 47.8 (Neutral); MACD: -0.93 (Bearish); 50-day MA: 271.63; 200-day MA: 289.15 \\
\midrule
\ AAPL Sector Alteration (Jan 27, 2023) 
& market\_news 
& Q1 earnings on Feb 2nd; Technology sector showing strong recovery; NASDAQ up 11.4\% YTD 
& Q1 earnings on Feb 2nd; Technology sector struggling; NASDAQ down 3.2\% YTD \\
& sector\_performance 
& Technology: +4.2\%; Communication Services: +3.1\%; Consumer Discretionary: +2.5\% 
& Technology: -2.8\%; Communication Services: -1.7\%; Consumer Discretionary: -0.9\% \\
\bottomrule
\end{tabularx}
\end{table*}

\section{Counterfactual Scenario Examples}
\label{app:counterfactual}

This appendix provides detailed examples of counterfactual scenarios to illustrate our evaluation framework for detecting information leakage in large language model (LLM)-based financial prediction agents. Detailed perturbations are shown in Table~\ref{tab:integrated}. 

\subsection{Counterfactual Scenario Framework}
\label{app:methodology}

Our counterfactual scenario framework constructs alternative versions of historical market scenarios by systematically perturbing key elements of the market state $S_t = \{P_t, F_t, N_t\}$, where $P_t \in \mathbb{R}^d$ represents price data, $F_t \in \mathbb{R}^m$ denotes market factors (e.g., technical indicators, fundamental factors), and $N_t \in \mathbb{R}^n$ represents factorized news. Perturbations are applied to prices, factors, or news while preserving statistical properties, such as volatility or sentiment distribution. For each case, we present:
\begin{itemize}
    \item \textbf{Original Scenario}: The actual historical market data provided to the model.
    \item \textbf{Counterfactual Scenario}: A perturbed version with specific modifications.
\end{itemize}

\subsection{Case Study 1: Earnings Announcement Perturbation (NVDA, May 25, 2022)}
\label{app:case1}

This case illustrates a counterfactual scenario for NVIDIA (NVDA) around its Q1 earnings announcement on May 25, 2022. The scenario tests model sensitivity to earnings-related news by perturbing the reported financial performance and outlook. The original scenario includes actual historical market data, such as price movements, technical indicators, and positive earnings news. The counterfactual scenario modifies the earnings news to reflect disappointing revenue, weaker data center and gaming performance, and a cautious outlook, while keeping price data and technical indicators unchanged. This perturbation aims to evaluate how models respond to significant changes in fundamental information while preserving other market signals.

\subsection{Case Study 2: Market Trend Reversal (TSLA, October 19, 2022)}
\label{app:case2}

This case illustrates a counterfactual scenario for Tesla (TSLA) around its Q3 2022 earnings announcement on October 19, 2022. The scenario tests model sensitivity to recent price trend information by reversing the 5-day price movement and adjusting related technical indicators. The original scenario includes actual historical market data, such as an upward price trend, bullish technical indicators, and earnings-related news. The counterfactual scenario reverses the 5-day price movement to reflect a downward trend, adjusts the RSI, MACD and other indicators to indicate bearish momentum, and keeps other data unchanged. This perturbation aims to evaluate how models respond to significant changes in short-term price trends while preserving other market signals.

\subsection{Case Study 3: Sector Performance Alteration (AAPL, January 27, 2023)}
\label{app:case3}

This illustrative case demonstrates a designed counterfactual scenario for Apple (AAPL) on January 27, 2023. The scenario tests model sensitivity to the broader macroeconomic market context by altering the performance of the technology sector and its associated news. The original scenario includes actual historical market data, such as a consistent upward price trend, bullish technical indicators, positive technology sector performance, and relevant supportive news. The counterfactual scenario, by contrast, modifies the technology sector performance to reflect a pronounced decline, while adjusting related news to indicate more negative market sentiment (e.g., a marked NASDAQ downturn and disappointing Microsoft cloud growth), and keeps other data unchanged. This systematic perturbation aims to evaluate how models robustly respond to significant changes in sector-level market signals while still faithfully preserving the underlying company-specific data.

\twocolumn[{
    \begin{center}
    \includegraphics[width=0.99\textwidth]{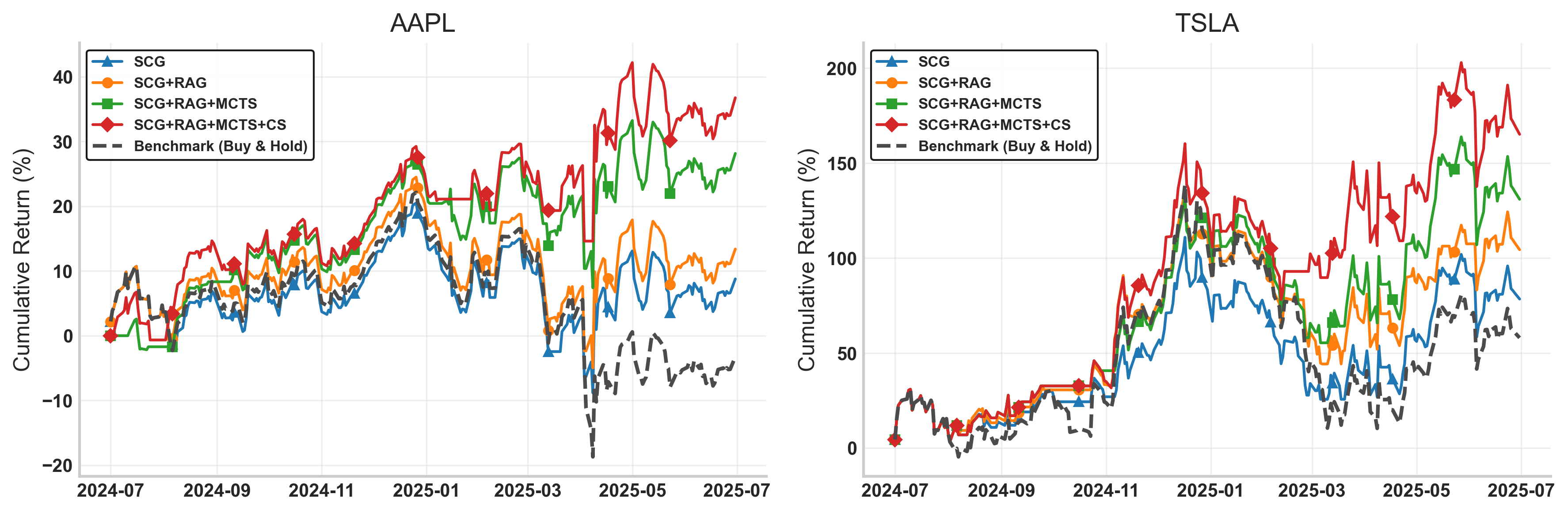}
    \captionof{figure}{Cumulative Return of FactFin with Different Components}
    \label{fig:crab}
    \vspace{0.7cm}
    \includegraphics[width=0.99\textwidth]{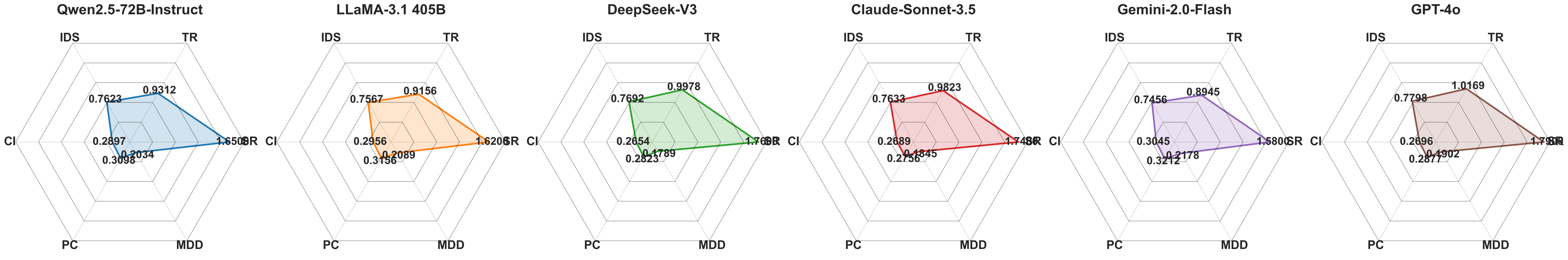}
    \captionof{figure}{LLM Backbone Performance}
    \label{fig:radar_chart}
    \end{center}
}]

\section{Effectiveness of Each Component}
\label{app:abl}

In Table~\ref{tab:ablation}, we study the effectiveness of the Strategy Code Generator (SCG), Retrieval-Augmented Generation (RAG), Monte Carlo Tree Search (MCTS), and Counterfactual Simulator (CS) in FactFin.
1) When using only SCG, FactFin exhibits low financial performance and high information leakage. Although SCG outperforms direct LLM decision-making, code generation alone is insufficient to effectively mitigate leakage. 
2) Adding RAG significantly improves TR and SR for both assets and reduces leakage, as structured market factor extraction enhances information utilization. 
3) Introducing MCTS further increases TR from 13.36\% to 28.12\% for AAPL and from 104.38\% to 130.93\% for TSLA, while reducing MDD from 23.65\% to 16.20\% for AAPL and from 37.45\% to 33.52\% for TSLA, substantially contribute to improved returns and risk reduction, with leakage moderately alleviated, though still at a high level.
4) The full model, including CS, achieves optimal performance across all metrics, with information leakage most effectively mitigated. The CS component is crucial for detecting and correcting leaks, allowing FactFin to avoid memorized patterns, ensuring strong performance and minimal data leakage.

The cumulative return trends for AAPL and TSLA across component combinations are shown in Figure~\ref{fig:crab}.

\section{Effectiveness of LLM Backbone}
\label{app:llm_comparison}

Table~\ref{tab:llm_comparison} comprehensively compares six representative LLMs as FactFin backbones. Overall, closed-source models consistently outperform their open-source counterparts in TR, SR, and leakage control, likely benefiting from broader training corpora and carefully engineered proprietary optimization techniques. GPT-4o demonstrates outstanding leadership in TR (101.69\%) and IDS (0.7798), while Claude-Sonnet-3.5 exhibits remarkable strength in PC (0.2756), thereby effectively minimizing undesirable pattern leakage. DeepSeek-V3, a competitive open-source model, delivers notably strong performance with the lowest MDD and CI, suggesting robust risk management capabilities and effective contextual leakage control. 

Encouragingly, all evaluated models achieve impressively high returns and relatively low leakage, thereby confirming FactFin’s robust adaptability across diverse LLM backbones. This observed consistency largely stems from FactFin’s carefully designed architecture, which strategically leverages components such as CS and MCTS to ensure stable and resilient performance. Notably, even open-source models can attain competitive outcomes when seamlessly integrated within FactFin, further highlighting the framework’s inherent capability to mitigate model-specific limitations. The comparative effectiveness of different LLM backbones is vividly illustrated in Figure~\ref{fig:radar_chart}, providing a comprehensive and intuitive visual overview of their nuanced trade-offs in return, risk, and leakage metrics.

\end{document}